\documentclass[12]{article}

\usepackage{arxiv}

\usepackage[utf8]{inputenc} 
\usepackage[T1]{fontenc}    
\usepackage{hyperref}       
\usepackage{url}            
\usepackage{booktabs}       
\usepackage{amsfonts}       
\usepackage{nicefrac}       
\usepackage{microtype}      
\usepackage{lipsum}

\usepackage{cite}
\usepackage{amsmath,amssymb,amsfonts}
\usepackage{algorithmic}
\usepackage{graphicx}
\usepackage{textcomp}
\usepackage{xcolor}
\graphicspath{{./Figures/}}
\def\BibTeX{{\rm B\kern-.05em{\sc i\kern-.025em b}\kern-.08em
    T\kern-.1667em\lower.7ex\hbox{E}\kern-.125emX}}

\title{Batch Recurrent Q-Learning for Backchannel Generation Towards Engaging Agents}
\author{
  Nusrah Hussain, Engin Erzin, T. Metin Sezgin, and Y\"{u}cel Yemez \\
  College of Engineering, Ko\c{c} University\\
   Istanbul, Turkey\\
  \textit{\{nhussain15, eerzin, mtsezgin,  yyemez\}@ku.edu.tr} \\
}

\begin{document}

    \maketitle

    \begin{abstract}
    The ability to generate appropriate verbal and non-verbal backchannels by an agent during human-robot interaction greatly enhances the interaction experience. Backchannels are particularly important in applications like tutoring and counseling, which require constant attention and engagement of the user. We present here a method for training a robot for backchannel generation during a human-robot interaction within the reinforcement learning (RL) framework, with the goal of maintaining high engagement level. Since online learning by interaction with a human is highly time-consuming and impractical, we take advantage of the recorded human-to-human dataset and approach our problem as a batch reinforcement learning problem. The dataset is utilized as a batch data acquired by some behavior policy. We perform experiments with laughs as a backchannel and train an agent with value-based techniques. In particular, we demonstrate the effectiveness of recurrent layers in the approximate value function for this problem, that boosts the performance in partially observable environments. With off-policy policy evaluation, it is shown that the RL agents are expected to produce more engagement than an agent trained from imitation learning. 
     
    \end{abstract}
    

\section{Introduction}
\label{intro}
Human-to-human interactions are characterized by a variety of verbal and non-verbal behaviors that work in conjunction
with dialog to make the interaction more engaging and impacting. Backchannels like non-verbal gestures (nods and smiles),
non-verbal vocalizations (mm, uh-huh, laughs) and verbal expressions (yes, right) play a significant role in enhancing
engagement and interest levels of the user \cite{turker2017analysis,inden2013timing}. Therefore, robots designed for domestic and interactive applications, also
need to have the intelligence of generating similar behaviors. In this work, we emphasize on backchannel generation targeted to maintain engagement of the user. Several definitions of engagement exist in the literature, which have been described in detail by Glas et al. \cite{glas2015definitions}. Poggi describes engagement as the value that a participant in an interaction attributes to the goal of being together with the other participant(s) and of continuing the interaction \cite{poggi2007mind}.

Generation of backchannels may be described as a sequential decision-making process during a dialog. These events are triggered based on some history of the interaction and they may impact the future course of the interaction. For example, a conversation may conclude prematurely in front of an unresponsive partner, while it may become very uncomfortable if the person in front laughs all the time. Though the objective behind backchannel generation by a human is quite complex, we chose to learn its optimization for increasing engagement of a user. This characteristic is particularly useful for applications like tutoring, companionship and instructive agents. The formulation of such a problem as a Markov decision process (MDP) comes naturally. If states are defined by a history of features (for example audio-visual), such that they follow the Markov property and a reward function is provided, a MDP may be structured. In our work, we experimented with reinforcement learning methods for learning the optimal policy for the MDP formulated by our problem. Since the state transition probability $p(s_{t+1}|s_t,a_t)$ is unknown, we deal with the class of model-free Q-learning based methods for continuous state and discrete action spaces.  

However, reinforcement learning comes with the need for an environment with which the agent may interact over many days as per training requirements. Such a facility cannot be achieved when the environment is a human. It will be highly time-consuming, intensely tiring and will demand much patience, especially when dealing with bad policies. There are numerous other real-world applications that face similar problem like robots in manufacturing, online advertisement and medical treatment recommendation systems where a bad policy may even be dangerous and illegal. Batch reinforcement learning techniques are expected to play a vital role in training RL agents for such environments. These techniques train an agent on a batch of samples gathered by a more controlled policy and have shown greater sample efficient since they may go over the collected data repeatedly. Inspired by this key advantage of the batch-RL and the availability of recorded human-to-human datasets on dyadic interactions, we approached our training as a batch-RL problem. We worked with the IEMOCAP dataset and processed it to define and extract tuples of the form $<s_t,a_t,r_t,s_{t+1}>$, which are required by batch-RL. Here $s_t$ is the present state, $a_t$ is action taken, $r_t$ is the reward received and $s_{t+1}$ is the next state the environment transitions to. Since the IEMOCAP was not designed to study engagement and does not represent near optimal demonstrations, instead of imitation the goal is to extract the best policy from the batch data. The success of the training was determined by Bellman residuals and the engagement expected from the learned policy.

The engineering of state definition in real-world problems is among the first challenge faced by researchers that may result in observations that do not reflect the true states. Thus, it is more useful to approach such problems as a partially observable Markov decision process (POMDP). Hausknecht et al. \cite{38hausknecht2015deep} present the advantage of recurrent layers like LSTMs when dealing with partial observability. In our formulation,we use speech features to define the current state of the user from the past 1 second time window. This definition may be considered as partially observable since there is a possibility that a true definition of the state needs to be formed by looking further into history. So we experimented with our definition of states using networks with and without LSTM layers. The recurrent layer improved the performance of our problem.

\section{Background}

\subsection{ Markov Decision Process}
In general, the reinforcement learning formulates the optimization problem as a Markov decision process (MDP) $(\mathcal{S,A,P,R,}\gamma)$ in which at each time step t, the environment observes a state $s_{t} \in \mathcal{S}$, the agent takes an action $a_{t} \in \mathcal{A}$ and the scalar reward $r_{t} \sim \mathcal{R}(s_t,a_t)$ is generated by the environment. Then the environment makes a transition to the next state $s_{t+1} \sim \mathcal{P}(s_t,a_t)$. The discount factor $\gamma \in [0,1)$ weighs the future rewards, determining the extent of temporal data that is affected by the current action. The solution to any MDP is a policy $\pi(a|s)$ which maximizes the expectation of sum of discounted rewards, i.e., the return. 

In real-world environments, it is not common for the true state information to be available. Instead, observations that hint about the underlying state may generally be received. That is, the Markov property may rarely hold. Therefore, approaching the problem as a partially observable MDP (POMDP) may present a greater advantage since it explicitly acknowledges that the observations are generated from a distribution of possible states. Formally, a POMDP can be described as the tuple $\mathcal{(S,A,P,R,O,}\Omega,\gamma)$. Here, $\mathcal{(S,A,P,\gamma)}$ are defined as earlier and additionally observation $o \in \Omega$ is generated from the probability distribution $o \sim O(s)$. 

\subsection{Q-learning}
Q-Learning \cite{watkins1992q} is a model-free off-policy algorithm for estimating the long-term expected return of executing an action from a given state. These estimated returns are known as Q-values. A higher Q-value indicates an action $a$ is judged to yield better long-term results in a state $s$. Q-learning follows the iterative process of fitting the Bellman control equation \cite{21sutton1998introduction}  given by:
\begin{equation}
    \label{eq:bellman}
    Q(s_t,a_t) = r_t+\gamma\max_{a'}Q(s_{t+1},a')
\end{equation}
So the loss function $L$ is defined by the mean square of the difference between right hand side and left hand side of the Bellman control equation. Deep Q-learning models the Q-values using neural networks, and hence the Q-function is represented by parameters $\theta$. The update equation is then given as:
\begin{equation}
    \label{eq:loss}
    \theta_{i+1} = \theta_i + \alpha \nabla_{\theta}L(\theta_i)
\end{equation}
Vanilla Deep Q-Learning has no explicit mechanisms for deciphering the underlying state of the POMDP and is only effective if the observations are reflective of underlying system states. In the general case, estimating a Q-value from an observation can be arbitrarily bad since $ Q(o,a| \theta ) \neq Q(s,a|\theta)$. The deep recurrent Q-learning architecture by \cite{38hausknecht2015deep} shows that adding recurrency allows the Q-network to better estimate the underlying system state, narrowing the gap between $Q(o,a|\theta)$ and $Q(s,a|\theta)$.

\subsection{Batch Reinforcement Learning}
Reinforcement learning (RL) algorithms generally fall into the category of online algorithms. As the agent interacts with the environment, it updates its policy towards high rewarding actions. However, this also means that the agent forgets its past experiences and cannot re-utilize the data from the state regions that were visited earlier. Batch reinforcement methods encourage first the collection of data and then learning of the policy from this batch data in an offline manner. This may be repeated several times but a pure batch reinforcement learning method performs one step of data collection followed by one step of offline learning. A more detailed survey on batch reinforcement learning can be found in \cite{lange2012batch}. Experience replay is a similar concept which initializes a fixed capacity of a buffer and keeps pushing new samples into the buffer and popping the old ones. Deep Q-network (DQN) \cite{mnih2015human} has shown that sampling randomly from the buffer to make updates, and hence breaking the sequential correlation between the samples, improves performance on Atari games. Another advantage of batch RL, that is of more interest to us, is that the batch data may be collected by any behavior policy, that may even be random. Several works exist on algorithms for batch reinforcement learning. The fitted Q-iterations (FQI) \cite{fqi} and neural fitted Q-learning (NFQ) \cite{NFQ} are among the more popular algorithms. The former uses a tree-based approach to model a Q-network while later modeled the Q-network with multi-layer perceptrons and fitting the Bellman optimality equation.

\section{Recent Work}
\subsection{Social Robots and RL}
Reinforcement learning has shown much success in a variety of domains and is a trending technique in the field of robotics. Several works have shown its use in training of an agent for behaviors similar to that of humans. The works by Qureshi et al. \cite{qureshi2016robot} \cite{qureshi2017show} presented an RL method for training an agent to greet as humans with the sequential actions of wait, look, wave and shake hand. They used multi-modal DQN and generated rewards at every successful handshake. The robot was trained for 14 days while it interacted with humans. In the work by Mitsunaga et al. \cite{mitsunaga2006robot}, RL is employed to adjust motion speed, timing, interaction distances, and gaze in the context of human-robot interaction (HRI). The reward is based on the amount of movement of the subject and the time spent gazing at the robot in one interaction. Recurrent neural networks were used in combination with Q-learning by Lathuili{\`e}re et al. \cite{lathuiliere2019neural} to find an optimal policy for robot gaze control in HRI. The training was performed in a simulated environment. In all these works, however, the agent either interacts with the environment (humans) for several days or training is done using simulators. We address the challenge where experience on a real physical system may be tedious to obtain, expensive, time-consuming and hard to simulate. We propose to use human-to-human interaction datasets as a batch of off-policy samples (trajectories) and use them in the context of offline batch reinforcement learning.

\subsection{Engagement in Interactions}
\label{related_work_engagement}
Poggi \cite{poggi2007mind} describes engagement as ``the value that a participant in an interaction attributes to the goal of being together with the other participant(s) and of continuing the interaction". An agent trained to maintain the engagement of a user is vital for several applications like companionship,  tutoring,  and  ambient assisting living.  A survey by Clavel et al. summarizes the issues regarding engagement in human-agent interactions, emphasizing its importance and indicating the growing interest of researchers in the field \cite{clavel2016fostering}. Verbal and non-verbal backchannels like nods, head tilts, eye-gaze, ‘hmms’ etc. are an important aspect of engagement and have been shown to promote engagement and interest levels of the user \cite{turker2017analysis,inden2013timing}. Researchers have mainly focused on rule-based backchannel generation \cite{al2009generating,liu2012generation} or data-driven unsupervised methods \cite{admoni2014data}. In this work, we show how to formulate the problem in a reinforcement learning framework and train an agent to learn a policy for backchannel generation that maximizes the engagement of the user.

One of the pioneering studies on the measurement of engagement is the work by Rich et al. \cite{rich2010recognizing}, where the authors propose an engagement model for collaborative interactions between human and computer.  They define four types of events as engagement indicators, referred to as connection events (CEs), which include directed gaze, mutual facial gaze, adjacency pair, and backchannels. Directed gaze event is defined when both participants look at a nearby object related to the interaction at the same time. The mutual facial gaze occurs when there is face-to-face eye contact. Adjacency pair indicates a successful event when turn taking occurs with some minimal time gap. Finally, backchannels refer to the generation of audio-visual feedback by a listener during the speaker's turn. In our work, we use these connection events to quantify engagement and generate a single scalar value at each time step to represent the rewards. An alternative option may be to directly annotate the engagement levels in the dataset. However, automatic detection of engagement allows the refinement of the policy in the future by continuously updating the policy as the agent interacts with humans.

\section{Proposed Methodology}
As described earlier, we propose a method to train an agent for the generation of backchannels that may maximize engagement using datasets as batch data. We work with the IEMOCAP (interactive emotional dyadic motion capture) dataset \cite{iemocap} that consists of dyadic human-to-human conversations on a range of scenarios. A total of 151 dialogues were performed by 10 professional actors in pairs on scripted and improvised scenes. In order to treat this as a batch data, we assume that of the two actors, one represents a behavior policy which takes the actions and the second actor behaves as an environment that generates states and rewards. Thus the IEMOCAP dataset may be viewed as a batch of trajectories collected by the behavioral policy. In order to double the training data, we also switch the roles of the actor as the behavior policy and the environment. Thus in total, we have 302 sequential decision making trajectories. 

\subsection{Batch-RL Formulaion}
Batch reinforcement learning algorithms work with tuples of the form $\langle s_t,a_t,r_t,s_{t+1} \rangle$ for $t=1:T$. At time $t$, $s_t$ is the state of the environment, $a_t$ is the action taken by the agent and $r_t$ is the reward. We extract these tuples at a rate of 40~Hz from the dataset, hence a batch data of approximately 3 million tuples is  produced. But the first step is to lay down the definitions of states, actions and rewards. Though our framework is general for any backchannel event, we perform experiments with laughs as a backchannel that may enhance engagement. States, actions, and rewards are defined as follows:
\begin{itemize}
	\item \textbf{State:} The state of the environment is represented by speech features extracted from past one second of data at every 25~msec step. This produces state information at a rate of 40~Hz. 
	\item \textbf{Action:} Agent's action is a binary variable, indicating the absence or presence of laugh. Laughs of the actor described as the behavioral policy are labeled at a rate of 40~Hz.
	\item \textbf{Reward:} The reward is a scalar quantity which comes from the engagement measures of the user at every time step. Engagement is calculated by determining the number of connection events in a time window. It is further elaborated below.
\end{itemize}

\begin{figure}[t]
	\begin{center}
		\centerline{\includegraphics[width=10cm,height=10cm,keepaspectratio]{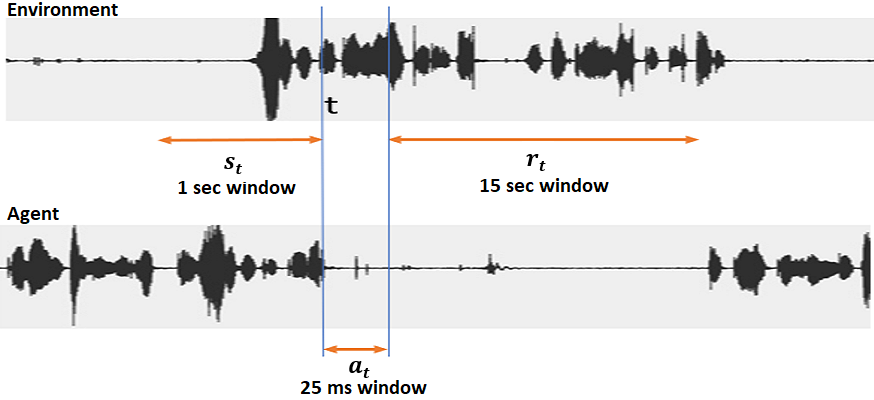}}
		\caption{Reinforcement learning formulation of speech driven backchannel generation (not drawn to scale)}
		\label{RLformulation}
	\end{center}
	\vskip -0.2in
\end{figure}

The states are defined using the mel-frequency cepstrum coefficients (MFCCs) and prosody features extracted from the speech signal of the environment. 13-dimensional MFCC features are computed using 40 milliseconds sliding Hamming window at intervals of 25 milliseconds. The speech intensity, pitch, and confidence-to-pitch with their first derivates make up a 6-dimensional prosody feature, so a 19-dimensional feature vector is formed when MFCCs and prosody are concatenated as in \cite{Bozkurt2016}. Following this, feature summarization is performed where a set of statistical quantities are computed that describe the short-term distribution of each feature over the past one second. These quantities comprise eleven functions, more specifically mean, standard deviation, skewness, kurtosis, range, minimum, maximum, first quantile, third quantile, median quantile and inter-quartile range, which were successfully used before by  \cite{metallinou2013tracking}. The dimension of each of these statistical feature vectors is 11 times the dimension of the corresponding feature vector. This makes the feature size of length 209.
 Fig.~\ref{RLformulation} shows the time windows used to extract states, rewards, and actions in one tuple. The dataset is pre-processed and such tuples are saved in a buffer.

Our measure of engagement is based on the method proposed in \cite{rich2010recognizing}, which is applicable to face-to-face collaborative HCI scenarios. Similar to the description in Section \ref{related_work_engagement}, we use the connection events (CE) (1) mutual facial gaze, (2) adjacency pair and (3) backchannels (that include laughs, smiles, nods and head-shakes) to quantify engagement. In \cite{rich2010recognizing}, the directed gaze event is defined when the agent and the participant look at a nearby object related to the interaction at the same time. However, in our dataset, since we do not have objects of interest at which both parties look at, we exclude it in our definition. The extracted CEs are then used to calculate a summarizing engagement metric called mean time between connection events (MTBCE). MTBCE measures the frequency of successful connection events that is for a given time interval T, MTBCE is calculated by T / (no. of CEs in T). As MTBCE is inversely proportional to engagement, similarly to \cite{rich2010recognizing}, we use pace = 1/MTBCE to quantify the engagement between a participant and the robot. The pace measure is calculated over a window of 15~seconds in our experiments.

\subsection{Q Networks}
\label{networks}
We perform experiments with two different neural networks and evaluate their relative performance. The first function approximation network is modeled as a multi-layer perceptron (MLP) to solve for a MDP where states follow the Markov property. It consists of 209 inputs (state feature size), two hidden layers with 100 and 25 neurons respectively and 2 outputs (Q-values for the two actions). For all the neurons ReLU activation function is used. For the second neural network, we introduce a LSTM layer to handle the problem as a POMDP. The same MLP structure is used. Only its first fully connected (FC) layer is replaced with a LSTM with identical number of neurons (i.e. 100). We also experiment with an LSTM replacing the second FC layer instead. However, since it did perform as well, we include here only the results for the case where the first FC layer is replaced.    

\section{Experiments}
For all our experiments the batch data is split into train and test sets in the ratio 4:1, hence 5-fold training is performed. It is split as leave one subject out (LOSO) and hence the reported results are subject independent. In all the experiments, the optimization is performed using Adam optimizer and a discount factor of value 0.99 is used. We train the two networks described in Section \ref{networks}: multi-layer perceptron (MLP) and fully connected LSTM (FC-LSTM). Following the concept of experience replay and randomization proposed by DQN \cite{mnih2015human}, the batch data is shuffled prior to training of the MLP network. In our second experiment,  the introduction of LSTM means the tuples cannot be randomized because the network now requires sequential data input. However, if each dialog is passed sequentially, the LSTM faces stability problem due to the long lengths of the sequences and the randomization proposed by DQN cannot be incorporated. We approach this by selecting randomly a starting position in each dialog and choose only the next $L$ time steps to pass to the network. We have tested with different sequence lengths $L$ and found $L=80$ (2 seconds of data) to be stable while improving the results. 

We have also tested with a linear function approximation network and found it to be unstable with diverging errors. Another comparison we have performed is with a policy learned from supervised learning. For supervised learning, we use the identical MLP network with a softmax layer at the end to produce probabilities of laugh and no laugh events. The loss is defined by the cross-entropy loss function with laugh labels as the true outputs. The value estimated by this policy is used as a baseline. 
\section{Results}
Evaluation of the resultant policy is a challenging problem since the environment (i.e., the human participant) is not readily available in our case. Although it is possible to conduct experiments with human-robot interactions, it is desirable to first understand the policy's effectiveness using quantitative measures. We use the Bellman residual and off-policy evaluation (OPE) techniques to understand the effectiveness of each policy

\subsection{Bellman Residual}
For a Q-value function approximation network $Q_\theta$, the Bellman residual is defined as the difference between the two sides of a Bellman control equation \cite{baird1995residual}. A smaller residual error means that the learned policy is closer to the optimal policy and is a true Q-function since it follows the Bellman equation more closely. Similar to the work of \cite{fqi}, we compute the Bellman residual, $B_r$, over the entire batch data $\mathcal{B}$ as
\begin{equation}
\label{bellmaneq}
B_r=\frac{1}{|\mathcal{B}|}\sum_{ \mathcal{B}} ( Q_{\theta}(s_t,a_t) - [ r_t + \gamma*\underset{a \in A}{\max} Q_{\theta}(s_{t+1},a) ])^2 .
\end{equation} 

\begin{figure}[t]\centering
\vspace{-0.1cm}
   \begin{minipage}{0.48\textwidth}
     \begin{center}
        \centerline{\includegraphics[width=8cm,height=8cm,keepaspectratio]{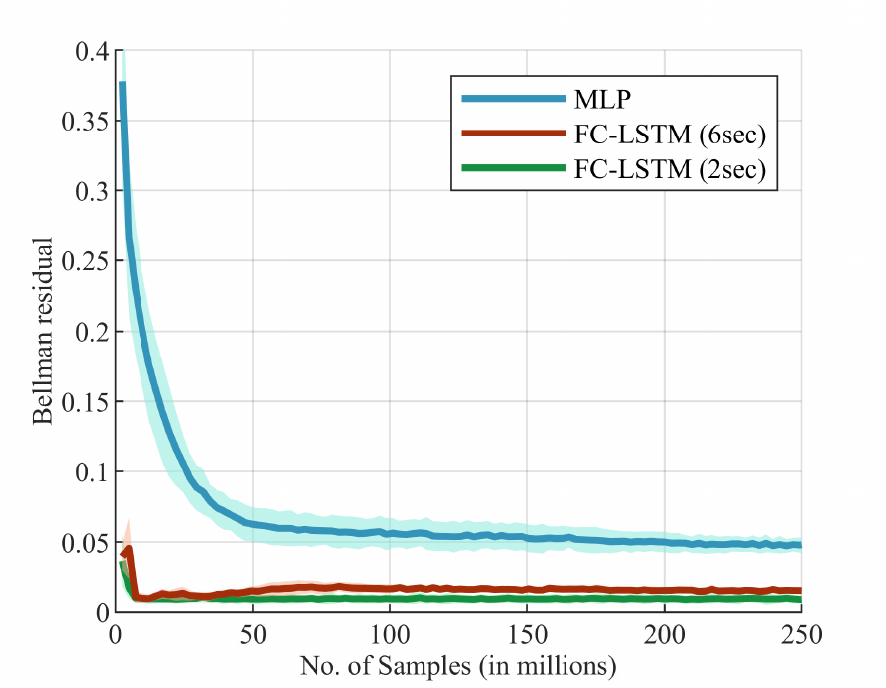}}
        \caption{Bellman residual vs training samples}\label{bellman_residual}
     \end{center}
   \end{minipage}
   \begin{minipage}{0.48\textwidth}
     \begin{center}
        \centerline{\includegraphics[width=8cm,height=8cm,keepaspectratio]{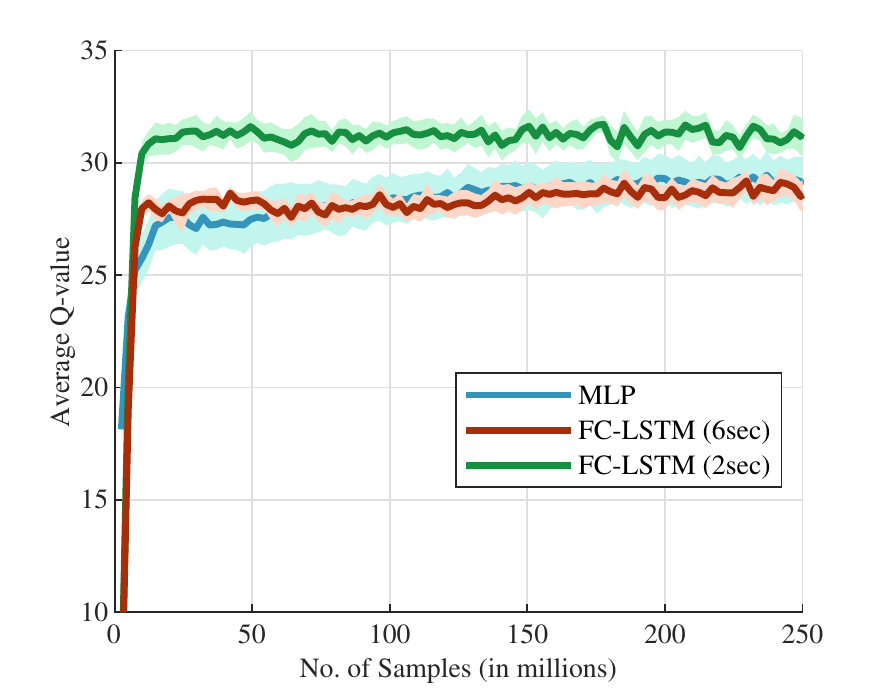}}
        \caption{Estimated Q-value vs training samples}\label{Qvalues}
     \end{center}
   \end{minipage}
\end{figure}

Fig. \ref{bellman_residual} shows the training curves for the multi-layer perceptron (MLP) and fully connected LSTM networks. Both models converge to a stable point. However, the lower Bellman residuals from the LSTM model show better optimality and validity of its value function approximation network. For comparison purpose, we also present the effect of different lengths of sequences used for LSTM training. Though truncation of sequences to $L$ samples may result in information loss from history, longer sequences become harder to train with LSTMs. The training curves for a length of past 2 seconds of data versus past 6 seconds of data are presented here. Fig. \ref{Qvalues} shows the trend of Q-values estimated by each network, averaged for the states present in the batch data. We observe that the estimates are close to each other, with LSTM network with 2 seconds of history surpassing marginally.

\subsection{Off-policy Policy Evaluation (OPE)} 
Off-policy policy evaluation (OPE) is used to predict the performance of policy with data only sampled by a behavior policy \cite{21sutton1998introduction}. In the last few years, many OPE techniques have emerged because of its importance in cases where a new policy cannot be tested directly with the environment \cite{thomas2016data,doroudi2017importance}. To compare the values of policies returned by each technique, we apply the step-wise weighted importance sampling (step-WIS) estimator given as
\begin{equation}
\label{eq:ope}
\hat{V}^{\pi}_{step-WIS}=\sum_{i=1}^n\sum_{t=0}^{T-1}\gamma^t\frac{\rho_t^{(i)}}{\sum_{i=1}^n \rho_t^{(i)}}r_t^{(i)},
\end{equation}
where $n$ is the number of trajectories, $T$ is the length of each trajectory and $\gamma$ is the discount factor. Then, the importance weight $\rho$ is defined as the ratio of the probability of the first $t+1$ steps of a trajectory under $\pi$ to the probability under a behavior policy $\pi_b$ and is given as $\rho_t = \prod_{i=0}^{t} \frac{\pi(a_i|s_i	)}{\pi_b(a_i|s_i)}$. The importance sampling approach to evaluation relies on using the importance weights $\rho_t$ to adjust for the difference between the probability of a trajectory under the behaviour policy $\pi_b$
and the probability under the evaluation policy $\pi$. Following the discussion of the work in \cite{raghu2018behaviour}, the behavior policy $\pi_b$ is estimated using approximate nearest neighbor \cite{Hyvonen2016}. Ideally  OPE needs  to  be  computed over  infinite  lengths,  but  taking into account  the  numerical  limitations, we calculate them over trajectories of length 200 samples with shifts at every sample.

\begin{figure}[!ht]
	\vspace{-0.1cm}
	\begin{center}
		\centerline{\includegraphics[width=8cm,height=8cm,keepaspectratio]{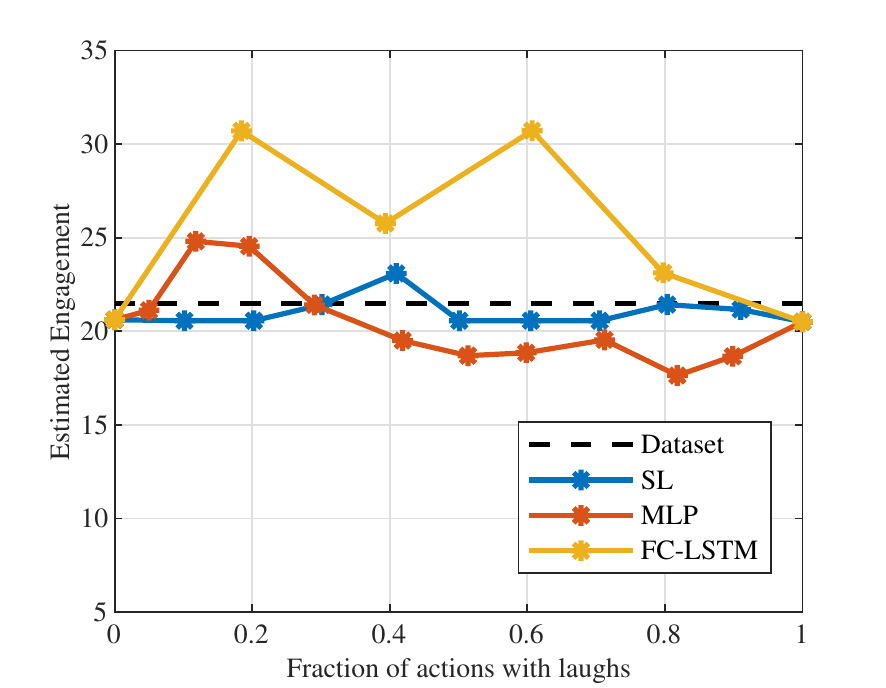}}
		\vspace{-0.1cm}
		\caption{Off-Policy Policy Evaluation}
		\label{fig:ope}
	\end{center}
	
\end{figure}

The training process for each technique produces an optimal Q-value function, which is used to implicitly define a policy. The simplest method is to act greedily and produce a deterministic policy by selecting the action with the highest Q-value. But since the OPE technique uses importance sampling and requires probabilities to reweigh each trajectory, we assign a high probability of $95\%$ to the action suggested by the greedy policy. Also, it is observed that each agent produces laughs much more frequently than the non-laughing event. On the contrary, in the dataset the laugh events occur only $\sim1.5\%$. In order to control the number of laughs, we use the softmax function to convert Q-values to probabilities and apply different thresholds to define a new deterministic policy. The OPE is performed at different threshold values (or amount of generated laughs) and similar to before a $95\%$ probability is assigned to the suggested action. Fig.~\ref{fig:ope} shows the values estimated by WIS at different fraction of laughs. The fully connected LSTM clearly outperforms MLP, producing a maximum value of 30.7 versus 24.8, whereas the batch data consists of an average value of 21.47. The policy from supervised learning is also tested in a similar fashion. Different thresholds are used to limit the number of laughs and OPE is performed from the deterministic actions suggested by the policy.  

\section{Conclusion and Future Work}
We demonstrated in this work batch reinforcement learning methods for training an agent to learn to produce backchannels with the objective of maximizing the user’s engagement.State modeling proved challenging for this problem and in general, may be described as a partially observable Markov decision process. Using audio features for state modeling, we successfully trained a RL-agent with Q-learning methods and demonstrated the superiority of recurrent structures like LSTMs in solving this problem. We have presented here the success of our training using various objective metrics. The future research direction may involve experiments with richer definitions of state with features like visual markers, emotional content, word embedding, etc. Additionally, we hope to extend the evaluation by performing experiments with human subjects and verifying from the feedback received.

\bibliographystyle{IEEEtran}

\end{document}